\documentclass{article}
\pdfoutput=1

\usepackage{arxiv}
\usepackage[utf8]{inputenc}             
\usepackage[T1]{fontenc}    		    
\usepackage{url}           			    
\usepackage[hidelinks]{hyperref}        
\usepackage{booktabs}       			
\usepackage{amsfonts}       			
\usepackage{nicefrac}       			
\usepackage{microtype}      			
\usepackage{graphicx}
\usepackage[sort&compress, numbers]{natbib}  
\usepackage{doi}
\usepackage{subfig}
\usepackage{placeins}  					
\usepackage{multirow}
\usepackage{amsmath}
\usepackage{amssymb} 
\usepackage{xcolor}

\newcommand{\MYhref}[3][black]{\href{#2}{\color{#1}{#3}}} 	

\hypersetup{
    colorlinks = true,
    linkbordercolor = white,
    linkcolor = blue,
    urlcolor  = blue,
	citecolor = blue,
}

\title{\texttt{rl\_reach}: Reproducible Reinforcement Learning Experiments for Robotic Reaching Tasks}

\author{\MYhref{https://orcid.org/0000-0002-9939-5537}{\includegraphics[scale=0.06]{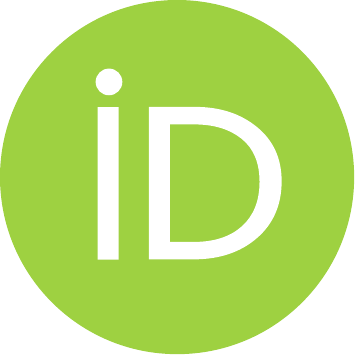}\hspace{1mm}Pierre Aumjaud}\thanks{Corresponding author} \\
	University College Dublin\\
	Dublin, Ireland \\
	\texttt{pierre.aumjaud@ucd.ie} \\
	\And
	David McAuliffe \\
	Resero Ltd \\
	Dublin, Ireland \\
	\texttt{david.mcauliffe@resero.io} \\
	\And
	\MYhref{https://orcid.org/0000-0002-8400-7079}{\includegraphics[scale=0.06]{Figures/orcid.pdf}\hspace{1mm}Francisco Javier Rodríguez Lera} \\
	Universidad de León\\
	León, Spain \\
	\texttt{fjrodl@unileon.es} \\
	\And
	\MYhref{https://orcid.org/0000-0002-4824-427X}{\includegraphics[scale=0.06]{Figures/orcid.pdf}\hspace{1mm}Philip Cardiff} \\
	University College Dublin\\
	Dublin, Ireland \\
	\texttt{philip.cardiff@ucd.ie} \\
}

\date{}

\renewcommand{\headeright}{}
\renewcommand{\undertitle}{}
\renewcommand{\shorttitle}{\texttt{rl\_reach}: Reproducible RL Experiments}

\hypersetup{
pdftitle={\texttt{rl\_reach}: Reproducible Reinforcement Learning Experiments for Robotic Reaching Tasks},
pdfsubject={cs.LG, stat.ML},
pdfauthor={Pierre Aumjaud, David McAuliffe, Francisco Javier Rodríguez Lera, Philip Cardiff}, 
pdfkeywords={Reinforcement Learning, Robotics, Benchmark, Model-free, Stable Baselines},
}

\begin{document}
\maketitle

\begin{abstract}
Training reinforcement learning agents at solving a given task is highly dependent on identifying optimal sets of hyperparameters and selecting suitable environment input / output configurations. This tedious process could be eased with a straightforward toolbox allowing its user to quickly compare different training parameter sets. We present \texttt{rl\_reach}, a self-contained, open-source and easy-to-use software package designed to run reproducible reinforcement learning experiments for customisable robotic reaching tasks. \texttt{rl\_reach} packs together training environments, agents, hyperparameter optimisation tools and policy evaluation scripts, allowing its users to quickly investigate and identify optimal training configurations. \texttt{rl\_reach} is publicly available at this URL: \href{https://github.com/PierreExeter/rl_reach}{https://github.com/PierreExeter/rl\_reach}.  
\end{abstract}


\begin{table}[!htbp]
\begin{tabular}{ll}
\toprule
Current code version & v1.0 \\
Permanent link to code/repository & \url{https://github.com/PierreExeter/rl_reach} \\
Permanent link to Reproducible Capsule & \url{https://codeocean.com/capsule/4112840/tree/} \\
Legal Code License   & MIT License \\
Code versioning system used & git \\
Software code language used & Python 3 \\
Compilation requirements \& dependencies & Docker OR Python 3, Conda, CUDA (optional)\\
Link to developer documentation/manual & \url{https://rl-reach.readthedocs.io/en/latest/index.html} \\
Support email for questions & \href{mailto:pierre.aumjaud@ucd.ie}{pierre.aumjaud@ucd.ie}\\
\bottomrule
\end{tabular}
\caption{Code metadata}
\label{tab:metadata} 
\end{table}

\FloatBarrier

\section{Context and Motivations}

Industrial processes have seen their productivity and efficiency increase considerably in recent decades thanks to the automation of repetitive tasks, notably with the advances in robotics. This productivity can be further improved by enabling robotic agents to solve tasks independently, without being explicitly programmed by humans.

Reinforcement Learning (RL) is a general framework for solving sequential decision-making tasks through self-learning and as such, it has found natural applications in robotics. In RL, an agent interacts with an environment by sending actions and receiving an observation -- describing the current state of the world -- and a reward -- describing the quality of the action taken. The agent's objective is to maximise the expected cumulative return by learning a policy that will select the appropriate actions in each situation.

RL has found many successful applications, however, experiments are notoriously hard to reproduce as the learning process is highly dependent on weight initialisation and environment stochasticity \cite{Henderson2018}. In order to improve reproducibility and compare RL solutions objectively, various standard toy problems have been implemented \cite{brockman2016, Bellemare2013, Beattie2016a, Nichol2018, Cobbe2019, Osband2019a}. A number of software suites provide training environments for continuous control tasks in robotics, such as \texttt{dm\_control} \cite{Tassa2018, Tassa2020}, Meta-World \cite{Yu2019a}, SURREAL \cite{Fan2018}, RLBench \cite{James2019a}, D4RL \cite{Fu2020}, robosuite \cite{Zhu2020} and robo-gym \cite{Lucchi2020}. 

We introduce \texttt{rl\_reach}, a self-contained, open-source and easy-to-use software package for running reproducible RL experiments applied to robotic reaching tasks. Its objective is to allow researchers to quickly investigate and identify promising sets of training parameters for a given task. \texttt{rl\_reach} is built on top of Stable Baselines 3 \cite{stable-baselines3} -- a popular RL framework. The training environments are based on the WidowX MK-II robotic arm and are adapted from the Replab project \cite{Yang2019}, a benchmark platform for running RL robotics experiments. \texttt{rl\_reach} encapsulates all the necessary elements for producing a robust performance benchmark of RL solutions for simple robotics reaching tasks. We aim to promote reproducible experimentation practice in RL research.

\section{Functionalities and Key Features}

\begin{figure*}[!htbp]
\centering
\includegraphics[width=0.7\linewidth]{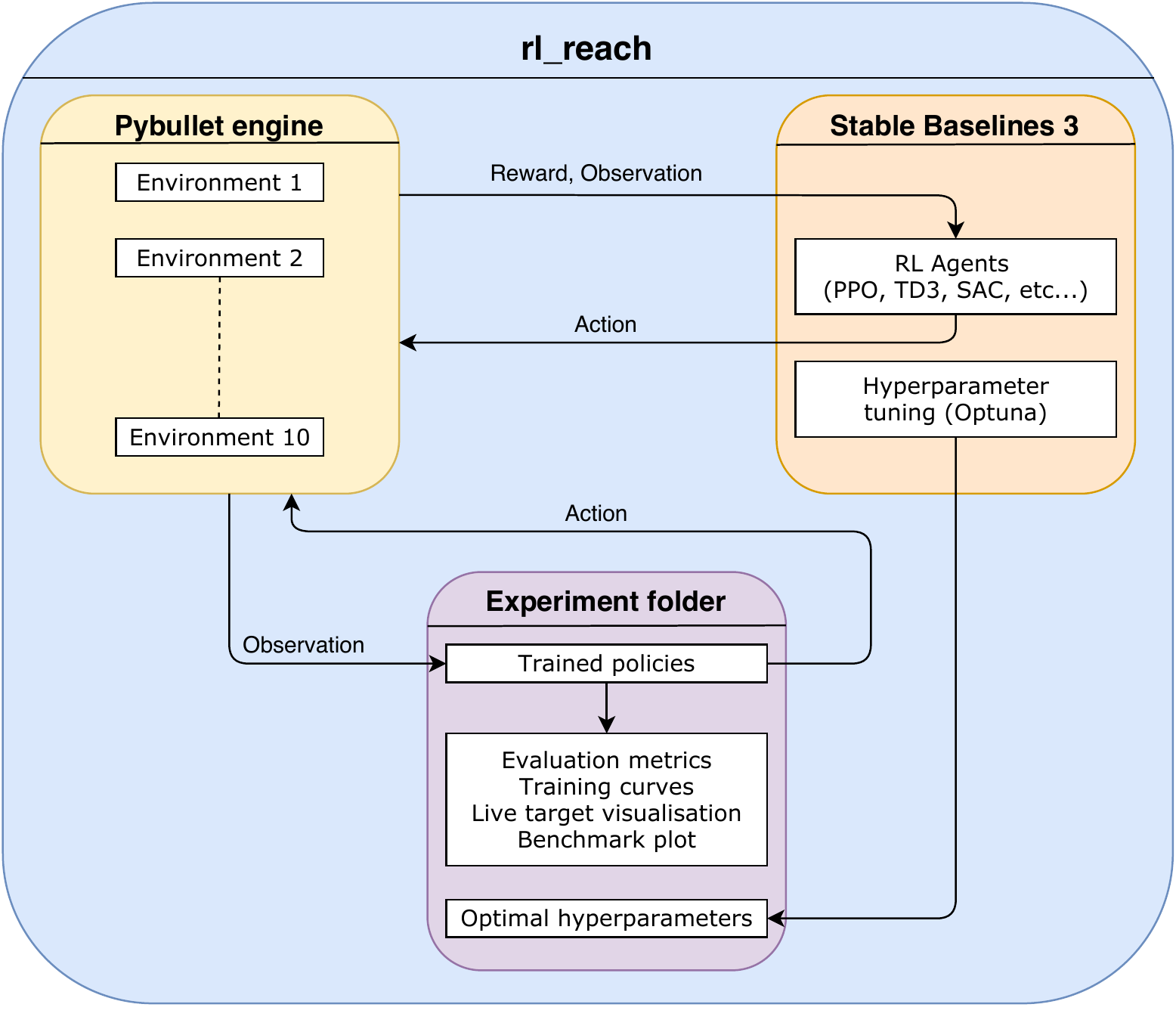}
\caption{\texttt{rl\_reach}'s flowchart and components}
\label{fig:diagram_rl_reach}
\end{figure*}

\begin{figure*}[!htbp]
\centering
\includegraphics[width=0.7\linewidth]{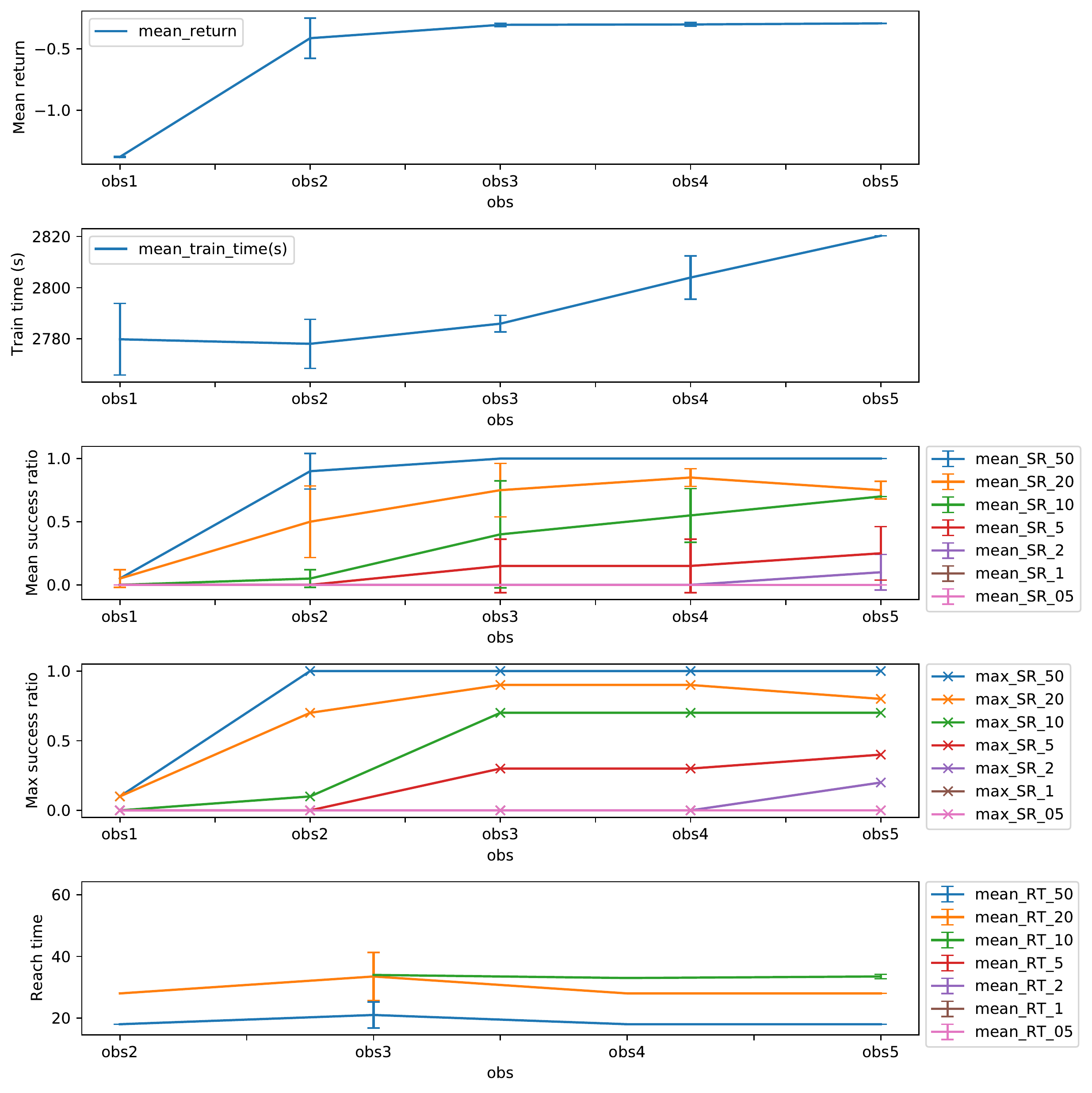}
\caption{An example of visualisation plot that compares the performance of different RL experiments}
\label{fig:benchmark_plot}
\end{figure*}

\begin{figure*}[!htbp]
\centering
\includegraphics[width=0.7\linewidth]{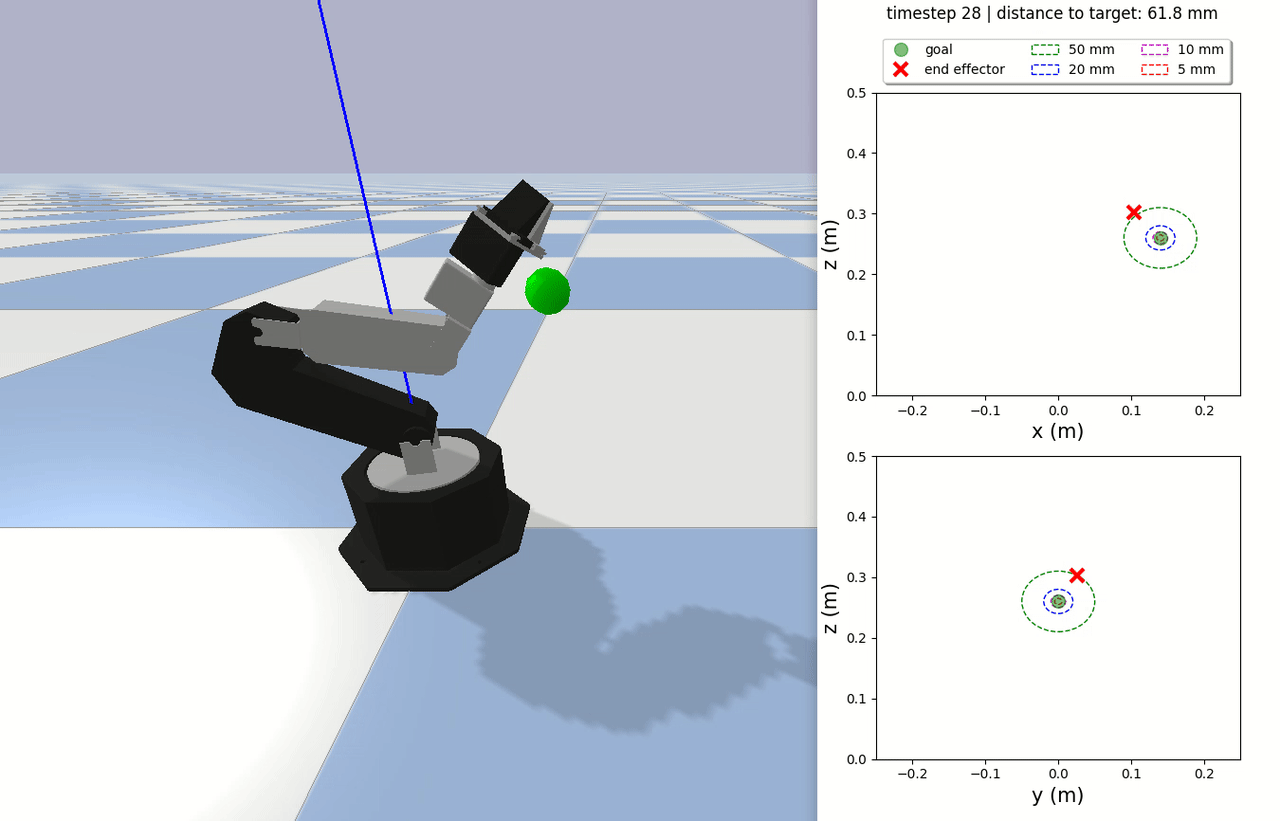}
\caption{The training environment with live visualisation of the end-effector and target position}
\label{fig:live_rendering}
\end{figure*}

\begin{figure*}[!htbp]
\centering
\includegraphics[width=\linewidth]{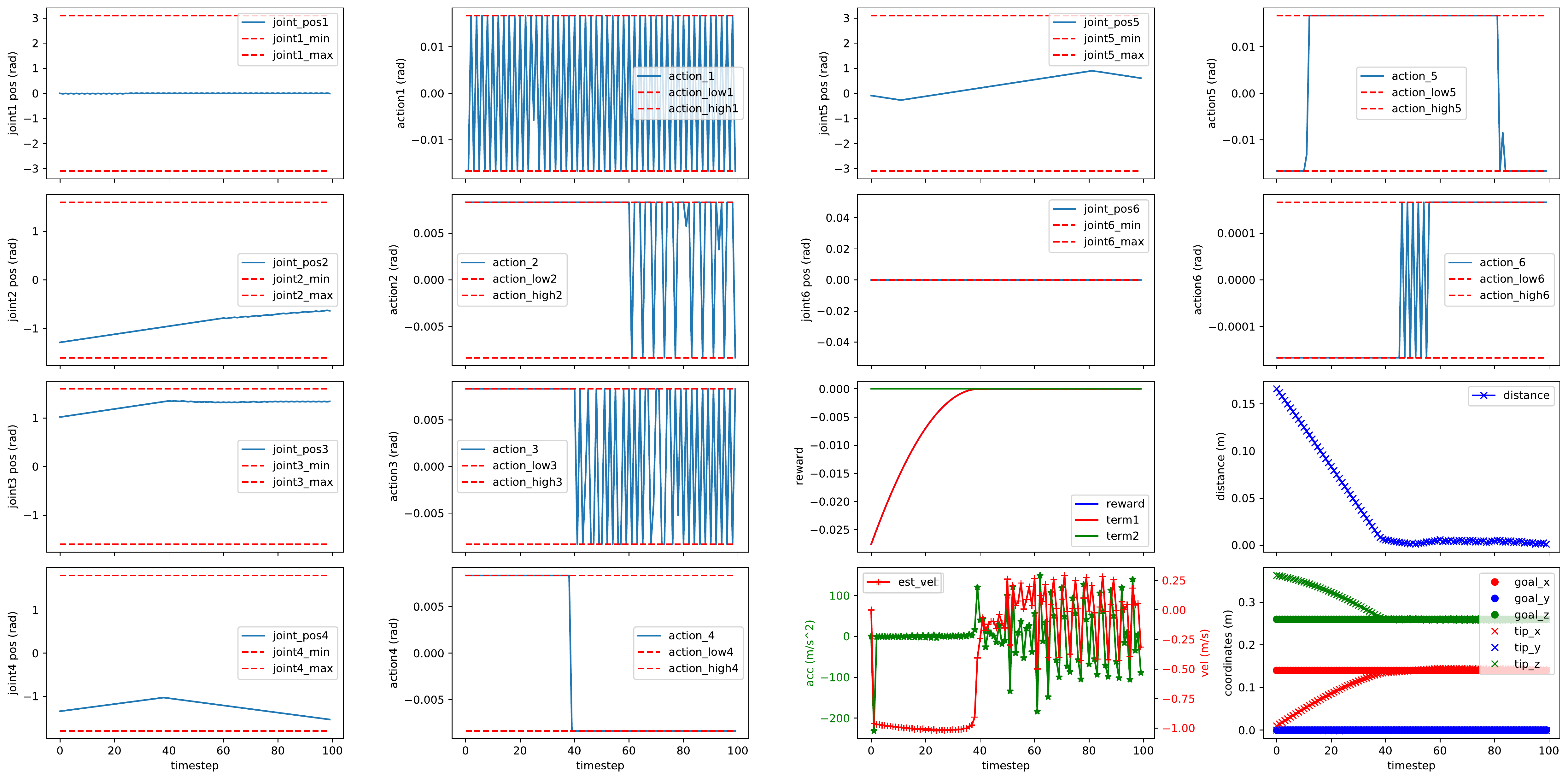}
\caption{An example of metadata plot after the evaluation of a trained policy}
\label{fig:plot_episode_eval_log}
\end{figure*}

\begin{figure*}[!htbp]
\centering
\includegraphics[width=0.7\linewidth]{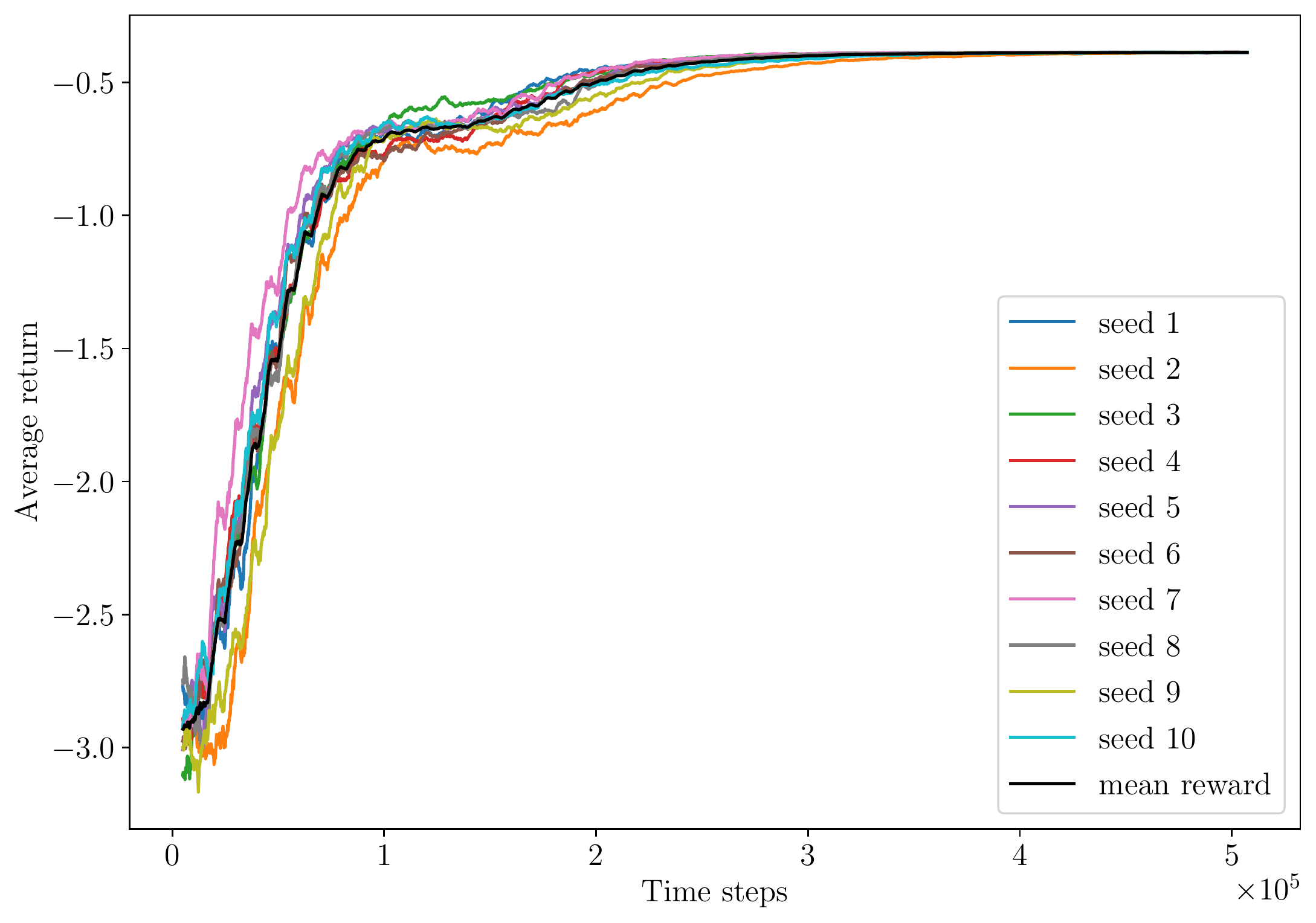}
\caption{An example of training curve plot}
\label{fig:reward_vs_timesteps_smoothed}
\end{figure*}

The \texttt{rl\_reach} software has been designed to quickly and reliably run RL experiments and compare the performance of trained RL agents against algorithms, hyperparameters and training environments. The code metadata are given in Table \ref{tab:metadata}. \texttt{rl\_reach}'s key features are:

\begin{itemize}
\item \textbf{Self-contained} : \texttt{rl\_reach} packs together a widely-used RL framework -- Stable Baselines 3 \cite{stable-baselines3}, training environments, evaluation and hyperparameter tuning scripts (Figure \ref{fig:diagram_rl_reach}). In addition to its ease of usability, only a few other packages offer such self-contained code.

\item \textbf{Free and open-source} : The source code is written in Python 3 and published under the permissive MIT license, with no commercial licensing restrictions. \texttt{rl\_reach} only makes use of free and open-source projects such as the deep learning library PyTorch \cite{Paszke2019} or the physics simulator Pybullet \cite{Coumans}. Many RL frameworks require a paid MuJoCo license, which can be an obstacle for sharing research results. Code quality and legibility is guaranteed with standard software development tools, including the Git version control system, Pylint syntax checker, Travis continuous integration service and automated tests.

\item \textbf{Easy-to-use} : A simple command-line interface is provided to train agents, evaluate policies, visualise the results and tune hyperparameters. Documentation is provided to assist end-users with the installation and main usage of \texttt{rl\_reach}. The software and its dependencies can be installed from source with the Github repository and Conda environment provided. Portability is maximised across platforms by providing \texttt{rl\_reach} as a Docker image, allowing it to run on any operating system that supports Docker. Finally, a reproducible code capsule is available online on the CodeOcean platform.

\item \textbf{Customisable training environments} : \texttt{rl\_reach} comes with a number of training environments for solving the reaching task with the WidowX robotic arm. These environments are easily customisable to experiment with different action, observation or reward functions. While many similar software packages exploit toy problems as benchmark tasks, \texttt{rl\_reach} provides its users with a training environment that is closer to an industrial problem, namely reaching a target position with a robotic arm.

\item \textbf{Stable Baselines inheritance} : Since \texttt{rl\_reach} is built on top of Stable Baselines 3 \cite{stable-baselines3} and its "Zoo", it comes with the same functionalities. In particular, it supports recent model-free RL algorithms such as A2C, DDPG, HER, PPO, SAC and TD3 and automatic hyperparameter tuning with the Optuna optimisation framework \cite{Akiba2019}.

\item \textbf{Reproducible experiments} : Each experiment (with a unique identification number) consists of a number of runs with identical training parameters but initialised with different initialisation seeds. The evaluation metrics are averaged across all the seed runs to promote reproducible, reliable and robust experiments.

\item \textbf{Straightforward benchmark} : When a trained policy is evaluated, the evaluation metrics, environment's variables and training hyperparameters are automatically logged in a CSV format. The performance of a selection of experiment runs can be visualised and compared graphically (Figure \ref{fig:benchmark_plot}).

\item \textbf{Debugging tools} : It is possible to produce a 2D or 3D live plot of the end-effector and goal positions during an evaluation episode (Figure \ref{fig:live_rendering}), as well as a number of physical characteristics of the environment such as the end-effector and the target position, the joint's angular position, reward, distance, velocity or acceleration between the end-effector and the target (Figure \ref{fig:plot_episode_eval_log}). It is also possible to plot the training curves for each individual seed run (Figure \ref{fig:reward_vs_timesteps_smoothed}). These plots have proven useful for debugging purposes, especially when testing a new training environment.
\end{itemize}

\section{Impact Overview}

Reinforcement Learning is a recent and highly active research field, with a relatively large number of RL solutions published every year. Accurately evaluating and objectively comparing novel and existing RL approaches is crucial to ensure continued progress in the field. Reproducing RL experimental results is often challenging due to stochasticity in the training process and training environments \cite{Henderson2018}. By providing a systematic tool for carrying out reproducible RL experiments, we hope that \texttt{rl\_reach} will promote better experimental practice in the RL research community and improve reporting and interpretation of results. Since \texttt{rl\_reach}'s interface is straightforward, intuitive and allows for a quick graphical comparison of experiments, it can be used as an educational platform for learning the practicalities of RL training.

Training RL agents is highly dependent on a number of intrinsic (e.g. initialisation seeds, reward functions, action shape, number of time steps) and extrinsic (algorithm hyperparameters) variables. Identifying the critical parameters that control a successful training can be a daunting task. Thanks to its easily customisable learning environments and extensive logging of training parameters, \texttt{rl\_reach} offers a unique solution to explore the effects of both intrinsic and extrinsic parameters on the training performance.

Finally, \texttt{rl\_reach} provides learning environments designed to train a robotic manipulator to reach a target position. This task is more industrially-relevant than many of the toy problems considered in other benchmark packages, thus allowing straightforward transfer of RL applications from academic research to industry.

A peer-reviewed article \cite{Aumjaud2021} has emanated from this software where the performance of robotics RL agents trained to reach target positions is compared. The trained policies were successfully transferred from the simulated to the physical robot environment.

\section{Conclusion and Potential Improvements}

We chose to focus on the reaching task as it is one of the simplest tasks to solve with a robotic arm, which allows users to run experiments with relatively low computing resources, while still being industrially relevant. Moreover, the reaching task allows the user to shape the reward easily and to implement training environments with both dense and sparse rewards. However, \texttt{rl\_reach} would benefit from supporting more complex and diverse manipulation tasks such as stacking, assembly, pushing or inserting. It also does not include the classic toy problems used traditionally for benchmarking RL agents. Finally, an implementation of the training environments for the physical WidowX arm would help validate the performance of policies trained in simulation.

\texttt{rl\_reach} has been designed as a self-contained tool, packaging both the training environments and the RL framework Stable Baselines 3 for convenience purposes. However this does not offer the flexibility to experiment with RL algorithms that are not supported by this framework. A potential future improvement would consist in producing a modular implementation of \texttt{rl\_reach} where both the training environments and the RL agents could be easily interchangeable.

\section*{Acknowledgements}

This Career-FIT project has received funding from the European Union’s Horizon 2020 research and innovation programme under the Marie Skłodowska-Curie grant agreement No. 713654.

\FloatBarrier

\bibliographystyle{spmpsci_unsrt}
\bibliography{library.bib}

\begin{thebibliography}{10}
\providecommand{\url}[1]{{#1}}
\providecommand{\urlprefix}{URL }
\expandafter\ifx\csname urlstyle\endcsname\relax
  \providecommand{\doi}[1]{DOI~\discretionary{}{}{}#1}\else
  \providecommand{\doi}{DOI~\discretionary{}{}{}\begingroup
  \urlstyle{rm}\Url}\fi

\bibitem{Henderson2018}
Henderson, P., Islam, R., Bachman, P., Pineau, J., Precup, D., Meger, D.: {Deep
  reinforcement learning that matters}.
\newblock 32nd AAAI Conference on Artificial Intelligence pp. 3207--3214 (2018)

\bibitem{brockman2016}
Brockman, G., Cheung, V., Pettersson, L., Schneider, J., Schulman, J., Tang,
  J., Zaremba, W.: {OpenAI Gym}.
\newblock arXiv \textbf{1606.01540} (2016)

\bibitem{Bellemare2013}
Bellemare, M.G., Veness, J.: {The Arcade Learning Environment : An Evaluation
  Platform for General Agents}.
\newblock Journal of Artificial Intelligence Research \textbf{47}, 253--279
  (2013)

\bibitem{Beattie2016a}
Beattie, C., Leibo, J.Z., Teplyashin, D., Ward, T., Wainwright, M., Lefrancq,
  A., Green, S., Sadik, A., Schrittwieser, J., Anderson, K., York, S., Cant,
  M., Cain, A., Bolton, A., Gaffney, S., King, H., Hassabis, D., Legg, S.,
  Petersen, S.: {DeepMind Lab}.
\newblock arXiv \textbf{1612.03801} (2016)

\bibitem{Nichol2018}
Nichol, A., Pfau, V., Hesse, C., Klimov, O., Schulman, J.: {Gotta Learn Fast: A
  New Benchmark for Generalization in RL}.
\newblock arXiv \textbf{1804.03720} (2018)

\bibitem{Cobbe2019}
Cobbe, K., Hesse, C., Hilton, J., Schulman, J.: {Leveraging Procedural
  Generation to Benchmark Reinforcement Learning}.
\newblock In: Proceedings of the 37th International Conference on Machine
  Learning, pp. 2048 -- 2056 (2020)

\bibitem{Osband2019a}
Osband, I., Doron, Y., Hessel, M., Aslanides, J., Sezener, E., Saraiva, A.,
  McKinney, K., Lattimore, T., Szepezvari, C., Singh, S., van Roy, B., Sutton,
  R., Silver, D., van Hasselt, H.: {Behaviour Suite for Reinforcement
  Learning}.
\newblock In: International Conference on Learning Representations (2020)

\bibitem{Tassa2018}
Tassa, Y., Doron, Y., Muldal, A., Erez, T., Li, Y., Casas, D.D.L., Budden, D.,
  Abdolmaleki, A., Merel, J., Lefrancq, A., Lillicrap, T., Riedmiller, M.,
  Benchmarking, F.: {DeepMind Control Suite}.
\newblock arXiv \textbf{1801.00690} (2018)

\bibitem{Tassa2020}
Tassa, Y., Tunyasuvunakool, S., Muldal, A., Doron, Y., Trochim, P., Liu, S.,
  Bohez, S., Merel, J., Erez, T., Lillicrap, T., Heess, N.: {dm{\_}control :
  Software and Tasks for Continuous Control}.
\newblock Software Impacts \textbf{6}(100022) (2020)

\bibitem{Yu2019a}
Yu, T., Quillen, D., He, Z., Julian, R., Hausman, K., Finn, C., Levine, S.:
  {Meta-World: A Benchmark and Evaluation for Multi-Task and Meta Reinforcement
  Learning}.
\newblock In: Conference on Robot Learning (CoRL) (2019)

\bibitem{Fan2018}
Fan, L., Zhu, Y., Zhu, J., Liu, Z., Zeng, O., Gupta, A., Creus-Costa, J.,
  Savarese, S., Fei-Fei, L.: {SURREAL: Open-Source Reinforcement Learning
  Framework and Robot Manipulation Benchmark}.
\newblock In: 2nd Conference on Robot Learning, \emph{Proceedings of Machine
  Learning Research}, vol.~87, pp. 767--782 (2018)

\bibitem{James2019a}
James, S., Ma, Z., Arrojo, D.R., Davison, A.J.: {RLBench: The Robot Learning
  Benchmark {\&} Learning Environment}.
\newblock IEEE Robotics and Automation Letters \textbf{5}(2), 3019--3026 (2020)

\bibitem{Fu2020}
Fu, J., Kumar, A., Nachum, O., Tucker, G., Levine, S.: {D4RL: Datasets for Deep
  Data-Driven Reinforcement Learning}.
\newblock arXiv \textbf{2004.07219} (2020)

\bibitem{Zhu2020}
Zhu, Y., Wong, J., Mandlekar, A., Mart{\'{i}}n-Mart{\'{i}}n, R.: {robosuite: A
  Modular Simulation Framework and Benchmark for Robot Learning}.
\newblock arXiv \textbf{2009.12293} (2020)

\bibitem{Lucchi2020}
Lucchi, M., Zindler, F., M{\"{u}}hlbacher-Karrer, S., Pichler, H.: {robo-gym --
  An Open Source Toolkit for Distributed Deep Reinforcement Learning on Real
  and Simulated Robots}.
\newblock In: IEEE/RSJ International Conference on Intelligent Robots and
  Systems (IROS) (2020)

\bibitem{stable-baselines3}
Raffin, A., Hill, A., Ernestus, M., Gleave, A., Kanervisto, A., Dormann, N.:
  {Stable Baselines3} (2019).
\newblock \urlprefix\url{https://github.com/DLR-RM/stable-baselines3}

\bibitem{Yang2019}
Yang, B., Zhang, J., Pong, V., Levine, S., Jayaraman, D.: {REPLAB: A
  Reproducible Low-Cost Arm Benchmark Platform for Robotic Learning}.
\newblock In: International Conference on Robotics and Automation (ICRA) (2019)

\bibitem{Paszke2019}
Paszke, A., Gross, S., Massa, F., Lerer, A., Bradbury, J., Chanan, G., Killeen,
  T., Lin, Z., Gimelshein, N., Antiga, L., Desmaison, A., K{\"{o}}pf, A., Yang,
  E., DeVito, Z., Raison, M., Tejani, A., Chilamkurthy, S., Steiner, B., Fang,
  L., Bai, J., Chintala, S.: {PyTorch: An imperative style, high-performance
  deep learning library}.
\newblock Advances in Neural Information Processing Systems \textbf{32},
  8026--8037 (2019)

\bibitem{Coumans}
Coumans, E., Bai, Y.: {PyBullet, a Python Module for Physics Simulation for
  Games, Robotics and Machine Learning} (2019).
\newblock \urlprefix\url{https://pybullet.org/}

\bibitem{Akiba2019}
Akiba, T., Sano, S., Yanase, T., Ohta, T., Koyama, M.: {Optuna: A
  Next-generation Hyperparameter Optimization Framework}.
\newblock In: Proceedings of the ACM SIGKDD International Conference on
  Knowledge Discovery and Data Mining, pp. 2623--2631 (2019).
\newblock \doi{10.1145/3292500.3330701}

\bibitem{Aumjaud2021}
Aumjaud, P., McAuliffe, D., Rodr{\'{i}}guez-Lera, F.J., Cardiff, P.:
  {Reinforcement Learning Experiments and Benchmark for Solving Robotic
  Reaching Tasks}.
\newblock In: Advances in Physical Agents II, pp. 318--331 (2021)

\end{thebibliography}

\end{document}